\documentclass[11pt]{article}

\usepackage[final]{acl}

\usepackage{latexsym}

\usepackage[expansion=false]{microtype}
\usepackage{inconsolata}

\usepackage{graphicx}
\usepackage{booktabs}
\usepackage{fontspec} 
\usepackage{newtxmath}
\usepackage[switch]{lineno}
\newfontfamily\hebrew{FreeSerif.otf}[
  Extension = .otf,
  Script=Hebrew
]
\usepackage{url}
\usepackage{multirow}
\usepackage{enumitem}
\usepackage{subcaption}

\title{Effective QA-driven Annotation of Predicate-Argument Relations \\ Across Languages}

\newcommand{\authorspace}{\hspace{9pt}}

\author{Jontahan Davidov$^{1}$ \authorspace Aviv Slobodkin$^1$ \authorspace Shmuel Tomi Klein$^1$ \authorspace  \\
\textbf{Reut Tsarfaty$^1$ \authorspace 
Ido Dagan$^1$ \authorspace Ayal Klein$^2$} \\
{$^1$Bar-Ilan University}\\{$^2$Ariel University}\\
 \footnotesize{\texttt{yonatand58@gmail.com}}
}

\begin{document}
\maketitle
\begin{abstract}
Explicit representations of predicate-argument relations form the basis of interpretable semantic analysis, supporting reasoning, generation, and evaluation.
However, attaining such semantic structures   requires costly annotation efforts and has remained largely confined to English.
We leverage the Question-Answer driven Semantic Role Labeling (QA-SRL) framework --- a natural-language formulation of predicate-argument relations --- as the foundation for extending semantic annotation to new languages.
To this end, we introduce a cross-linguistic projection approach that reuses an English QA-SRL parser within a constrained translation and word-alignment pipeline to automatically generate question-answer annotations aligned with target-language predicates.
Applied to Hebrew, Russian, and French --- spanning diverse language families --- the method yields structurally rich training data and fine-tuned, language-specific parsers that outperform strong multilingual LLM baselines (GPT-4o, LLaMA-Maverick).
By leveraging QA-SRL as a transferable natural-language interface for semantics, our approach enables efficient and broadly accessible predicate-argument parsing across languages. 
\end{abstract}

\section{Introduction}\label{sec_introduction}

Despite the representational power of large language models (LLMs), explicit predicate-argument representations remain a cornerstone of natural language understanding.
By decomposing sentences into elementary meaning units --- \textit{who did what to whom,  how, when} and \textit{where} --- such structures enable fine-grained modeling of meaning and support tasks that depend on precise semantic alignment across texts, including faithfulness and attribution analysis in generation, controlled text production, and systematic evaluation of generated content \citep{bhattacharyya-etal-2022-aligning, dryjanski-etal-2022-samsung, fan-etal-2023-evaluating, zhang2025qapyramid}.

Traditional semantic role labeling (SRL) frameworks such as FrameNet, PropBank and OntoNotes \citep{baker1998berkeley, kingsbury2002propbank, weischedel-etal-2013-ontonotes} provide such explicit representation of predicate-argument structure through predefined role inventories and frame lexicons --- an approach also adopted by later meaning representations such as AMR, UCCA, and other frameworks \citep{banarescu-etal-2013-abstract, abend2013ucca, oepen2015semeval}. 
However, these linguistically grounded schemas rely on expert annotation and language-specific resources, making large-scale annotation costly and, consequently, largely limited to English.
As a result, cross-lingual semantic annotation has progressed slowly and unevenly, despite the conceptual centrality of predicate-argument representations in NLP.

\begin{table*}[]
\resizebox{\textwidth}{!}{%
\begin{tabular}{|lrll|}
\hline
\multicolumn{4}{|l|}{Both were shot in the confrontation with police and have been recovering in hospital since the massive attack.} \\ \hline
\multicolumn{1}{|l|}{QA-SRL} & \multicolumn{1}{r|}{1}  & \multicolumn{1}{l|}{When was someone shot?}                          & in the confrontation ; the attack    \\
\multicolumn{1}{|l|}{}       & \multicolumn{1}{r|}{2}  & \multicolumn{1}{l|}{Who was shot?}                                   & Both                                 \\
\multicolumn{1}{|l|}{}       & \multicolumn{1}{r|}{3}  & \multicolumn{1}{l|}{Who shot someone?}                               & police                               \\
\multicolumn{1}{|l|}{}       & \multicolumn{1}{r|}{4}  & \multicolumn{1}{l|}{Where has someone been recovering?}              & in hospital                          \\
\multicolumn{1}{|l|}{}       & \multicolumn{1}{r|}{5}  & \multicolumn{1}{l|}{How long was someone recovering from something?} & since the attack                     \\
\multicolumn{1}{|l|}{}       & \multicolumn{1}{r|}{6}  & \multicolumn{1}{l|}{Who was recovering from something?}              & Both                                 \\
\multicolumn{1}{|l|}{}       & \multicolumn{1}{r|}{7}  & \multicolumn{1}{l|}{What was someone recovering from?}               & shot                                 \\ \hline
\multicolumn{1}{|l|}{QANom}  & \multicolumn{1}{r|}{8}  & \multicolumn{1}{l|}{Who confronted with something?}                  & Both                                 \\
\multicolumn{1}{|l|}{}       & \multicolumn{1}{r|}{9}  & \multicolumn{1}{l|}{What did someone confront with?}                 & police                               \\ \hline
\end{tabular}%
}
\caption{An example sentence annotated with QA-SRL and QANom.}
\label{tab:example_qasem}
\end{table*}

Question-Answer driven SRL (QA-SRL) \citep{he2015qasrl} introduced a natural-language alternative: representing predicate-argument relations through question-answer pairs rather than symbolic role labels.
Each predicate is associated with simple questions (e.g., \textit{Who \textbf{shot} someone?}) and their corresponding answer spans, capturing the underlying roles without schema-specific design (see Table \ref{tab:example_qasem} for a full example).
This formulation yields an interpretable, annotation-friendly, and LLM-compatible representation of semantic structure.
Subsequent work extended this approach to deverbal nominalizations  \citep[QANom;][]{klein-etal-2020-qanom} as well as to a broader QA-based semantic framework \citep{klein-etal-2022-qasem} encompassing additional types of predication \citep{pyatkin2020qadiscourse, pesahov-2013-qaadj}.
In practice, QA-SRL has been shown to be effective across multiple downstream tasks \citep{brook-weiss-etal-2021-qa, sultan-shahaf-2022-life, caciularu-etal-2023-peek, cattan2024localizing, zhang2025qapyramid}, underscoring its role as a broadly applicable representation of predicate-argument structure.

However, existing QA-SRL resources have so far been developed exclusively for English.
To realize its broader potential, we seek to leverage QA-SRL’s natural-language format as a vehicle for scaling predicate-argument annotations to new languages with minimal language-specific prerequisites. Achieving this requires a systematic, cross-lingual extension that preserves QA-SRL’s accessibility while maintaining the structured correspondence between predicates, questions, and answer spans that underpins its semantic fidelity.

In this work, we propose a multilingual QA-SRL projection approach that fulfills this goal.
Our algorithm reuses a high-quality English QA-SRL parser within a refined projection pipeline that translates the English question-answer annotations into the target language.
It combines constrained machine translation, word alignment, and QA-structure preservation mechanisms specifically tailored to QA-SRL’s semi-structured format.
The resulting projected annotations are used to supervise the fine-tuning of lightweight, language-specific QA-SRL parsers, making automatic predicate-argument analysis efficiently attainable and widely accessible across languages. 
To ground the task and illustrate the expected outputs across languages, Table \ref{tab:qa-examples} provides concrete examples of QA-SRL predictions for the same predicate in Hebrew, Russian, and French.

We validate this approach on three typologically diverse languages --- Hebrew, Russian, and French --- spanning Semitic, Slavic, and Romance families.
For each, we construct projected verbal and nominal QA-SRL datasets, curate gold evaluation subsets, and train language-specific QA parsers that substantially outperform strong multilingual LLM baselines such as GPT-4o and LLaMA-Maverick.

Taken together, our results establish a scalable methodology for extending predicate-argument semantics to new languages.
By operationalizing multilingual QA-SRL as a projection-based process, we demonstrate how explicit predicate-argument representations --- long central to semantic analysis --- can be scaled cross-linguistically with minimal cost, leveraging natural language itself as the medium of meaning transfer.

\section{Background and Related Work}
\label{sec_background}

\begin{table*}[h!]
\small
\centering
\renewcommand{\arraystretch}{1.2}
\begin{tabular}{|p{1.5cm}|p{2.8cm}|p{2.9cm}|p{2.4cm}|p{4.1cm}|}
\hline
\textbf{Language} & \textbf{Sentence} & \textbf{Question} & \textbf{Answer} & \textbf{English gloss} \\
\hline
Hebrew & {\scshape\scriptsize\hebrew{המדיניות את אישרה הוועדה }} {\scshape\scriptsize\hebrew{החדשה}} 
& {\scshape\scriptsize\hebrew{משהו? אישר מי}} 
& {\scshape\scriptsize\hebrew{הוועדה}} 
& Who approved something? → The committee \\
& & {\scshape\scriptsize\hebrew{אישר? מישהו מה}} 
& {\scshape\scriptsize\hebrew{החדשה המדיניות את}} 
& What did someone approve? → the new policy \\
\hline
Russian & {\footnotesize Комитет \textbf{одобрил} новую политику}
& {\footnotesize Кто что одобрил?} 
& {\footnotesize Комитет} 
& Who approved something? → The committee \\
& & {\footnotesize что кто-то одобрил?} 
& {\footnotesize новую политику} 
& What did someone approve? → the new policy \\
\hline
French & {\footnotesize Le comité a \textbf{approuvé} la nouvelle politique} 
& {\footnotesize qui a approuvé quelque chose?} 
& {\footnotesize Le comité} 
& Who approved something? → The committee \\
& & {\footnotesize qu'est-ce que quelqu'un a approuvé?} 
& {\footnotesize la nouvelle politique} 
& What did someone approve? → the new policy \\
\hline
\end{tabular}
\caption{Examples of model predictions in different target languages.}
\label{tab:qa-examples}
\end{table*}

This section situates our work in two strands of prior research: efforts to build multilingual semantic resources, and the development of QA-based representations.

\subsection{Multilingual Semantic Resources}

Large-scale semantic resources remain concentrated in English. While syntactic treebanks such as Universal Dependencies provide broad coverage \citep{nivre-etal-2016-universal-v1,de-marneffe-etal-2021-universal}, comparable semantic annotations such as predicate-argument relations are scarce beyond English, limiting interpretable, content-grounded modeling in most languages.

Cross-lingual SRL has relied on translating English corpora and projecting PropBank-style roles via alignments, or on inducing training data from parallel corpora. 
A prominent example is Universal Proposition Bank (UPB), which constructs multilingual propbanks through a two-stage pipeline combining monolingual SRL with multilingual parallel data, and has recently been extended and improved in UP2.0 \citep{jindal-etal-2022-universal}. 
Beyond projection, another core obstacle is \emph{formalism heterogeneity}: different languages adopt different SRL inventories (e.g., PropBank vs.\ AnCora vs.\ PDT-Vallex), making cross-lingual transfer depend on schema mapping. 
\citet{conia-etal-2021-unifying} address this by training a unified model over heterogeneous SRL resources that implicitly learns cross-inventory correspondences without relying on word alignment or translation.
In parallel, lexical-semantic resources such as VerbAtlas define cross-frame semantic roles and prototypical argument structures, offering an alternative to purely predicate-specific PropBank role sets and supporting more uniform role semantics \citep{di-fabio-etal-2019-verbatlas}.

Alongside these efforts, recent cross-lingual approaches include translated training data \citep{fei-etal-2020-cross}, X-SRL’s multilingual projection \citep{daza2020xsrlap}, alignment-free modeling \citep{cai-lapata-2020-alignment}, and divergence-aware corrections \citep{youm2024dahrs}. 
Similar strategies underlie AMR transfer: contextual word alignments \citep{sheth-etal-2021-bootstrapping} or translate-then-parse baselines \citep{uhrig-etal-2021-translate} achieve strong results, but again require adaptation to formalism-specific schemata. 
Closer to our direction is CLaP \citep{parekh-etal-2024-contextual}, which improves projection by conditioning label translation on context and label semantics rather than word-level alignments, yet still operates within a schema-based SRL framework. 
Recent surveys stress that reliance on English-centric linguistic formalisms, schema mapping, and costly alignment pipelines constrains scalability across languages \citep{hammerl-etal-2024-understanding}.

These limitations motivate exploring natural-language-based predicate-argument representations, which can mitigate reliance on rigid role-schema mappings and enable more scalable, cost-effective transfer across languages.



\subsection{QA-based Semantic Role Labeling}
QA-driven semantic role labeling offers an alternative to schema-based SRL by representing predicate--argument relations directly in natural language, rather than through fixed role inventories.
As illustrated in Table~\ref{tab:example_qasem}, each predicate is queried with simple questions capturing its roles, and the answers mark the corresponding argument spans.
This formulation eliminates the need for predefined role sets or cross-schema mapping, enables intuitive crowdsourced annotation, and naturally aligns with the capabilities of modern language models \citep{he2015qasrl}.

The QA-SRL framework was extended to include deverbal nominalizations \citep{klein-etal-2020-qanom}, preserving the same question formats and capturing eventive structure in nominal domains. 
Verbal and nominal QA-SRL have since been modeled jointly through a text-to-text framework \citep{klein-etal-2022-qasem}, with subsequent advances yielding a state-of-the-art English parser \citep[][see Appendix~\ref{app:QASem-parser} for details]{cattan2024localizing}. This parser provides the English annotations that we project in our cross-lingual pipeline.

While this work focus on verbal and nominal QA-SRL, which encode the core propositional content of sentences, other extensions of the QA-based semantic paradigm cover adjectival predicates \citep{pesahov-2013-qaadj}, discourse relations  \citep{pyatkin2020qadiscourse} and additional noun-related semantics \cite{tseytlin2025qanoun}. We leave the cross-linguistic extension of these for future exploration.

Recent studies have demonstrated the utility of QA-SRL across a range of downstream applications, including cross-text predicate-argument alignment \citep{brook-weiss-etal-2021-qa}, fine-grained summarization evaluation \citep{zhang2025qapyramid}, localization of factuality assessment \citep{cattan2024localizing}, event similarity in analogical reasoning \citep{sultan-shahaf-2022-life}, and multi-document pre-training \citep{caciularu-etal-2023-peek}.
These applications establish QA-SRL as a general-purpose, interpretable content representation at the fine-grained level of predicate-argument relations. 
Yet crucially, all prior work has been limited to English.  
Our contribution is to address this gap by scaling QA-SRL to new languages through a cross-lingual translation pipeline. 

\section{Multilingual QA-SRL Projection Algorithm}
\label{sec_qasem_translation}

\subsection{Overview}
\label{subsec:projection-overview}

\paragraph{Goal \& Motivation}
Our goal is to demonstrate a general methodology for attaining predicate-argument representations automatically and at low cost in new languages. Concretely, our methodology produces large-scale, high-quality QA-SRL training datasets in a target language~$\mathcal{L}$ for fine-tuning a language-specific QA-SRL parser. 
Our proposed projection pipeline, introduced in the current section, takes a target-language corpus and automatically generates question-answer annotations that serve as supervision for training these parsers, making the trained models the ultimate output of the process.

In contrast to traditional SRL projection approaches that transfer schema-bound role labels, our method projects natural language question-answer pairs, avoiding the need for role mapping and leveraging core translation capabilities. On the other hand, we observe that a naive English back-translation approach for attaining QA-SRL QAs in $\mathcal{L}$ yields inadequate QA-SRL outputs (see Appendix \ref{app:surface_translation_limitations} for a detailed discussion), which motivates our more refined projection approach.   

\paragraph{Target Language Requirements}
Our projection pipeline is applicable in any target language~$\mathcal{L}$ with the following broadly available resource requirements:
(i) a reasonably sized corpus with POS annotations to identify verbal and candidate nominal predicates; 
(ii) machine translation (MT) from $\mathcal{L}$ to English;
(iii) word alignment tools; and
(iv) a pretrained language model capable of simple in-context operations --- primarily constrained translation (\S \ref{subsec:constrained_translation}). 
In practice, these prerequisites are already satisfied for most commonly used languages, as contemporary multilingual resources and models provide dependable coverage for well over a hundred languages, encompassing the vast majority of the world’s population.\footnote{The
\href{https://universaldependencies.org/}{Universal Dependencies  project} (UD) currently provides treebanks for over 150 languages \citep{nivre-etal-2020-universal-v2}.
Major parallel-data and translation resources such as OPUS \citep{tiedemann2022open}, NLLB \citep{costa2022nllb}, and SeamlessM4T \citep{barrault2023seamlessm4t} collectively cover between 100 and 200 languages, while large-scale MT systems such as Google Translate now support around 250 languages (see \href{https://simple.wikipedia.org/wiki/Google_Translate}{Wikipedia}). Modern multilingual LLMs (e.g., GPT-4o) also exhibit strong zero-shot cross-lingual competence, further relaxing the need for language-specific models and annotations.} 


\paragraph{Projection Pipeline Outline} The pipeline operates in four stages (Figure~\ref{schema}):

\begin{figure}[t]
    \centering
    \includegraphics[width=\columnwidth, keepaspectratio]{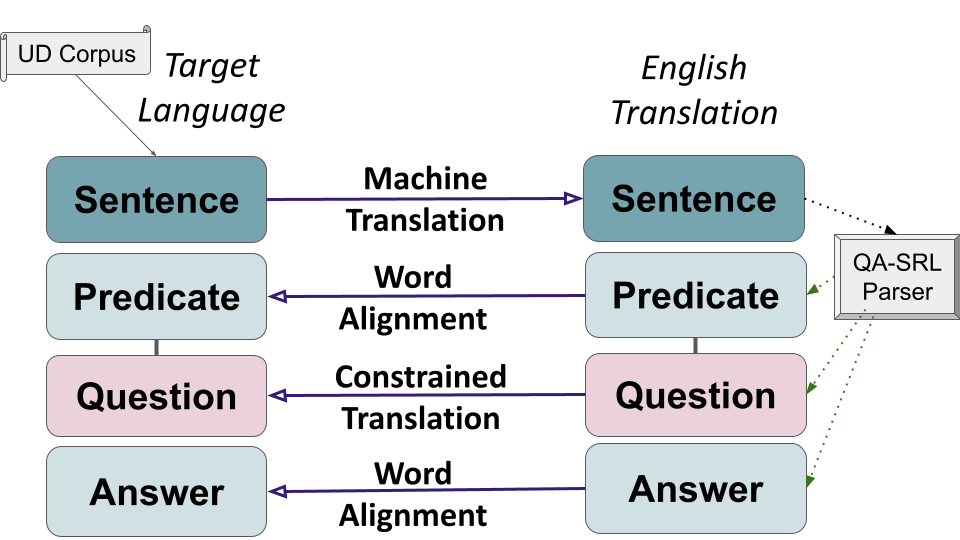}
    \caption{A schematic overview of our proposed methodology for transferring QA-SRL to new languages based on the English QA-SRL infrastructure.}
    \label{schema}
\end{figure}

\begin{enumerate}[noitemsep, topsep=0pt]
    \item \textbf{Sentence translation:} Starting from a tokenized, part-of-speech tagged corpus in $\mathcal{L}$, each sentence is translated into English.\footnote{We used open-source translation models from the \href{https://huggingface.co/Helsinki-NLP}{Helsinki-NLP} project for translating sentences from the target language into English: \href{https://huggingface.co/Helsinki-NLP/opus-mt-tc-big-he-en}{\texttt{opus-mt-tc-big-he-en}} for Hebrew, \href{https://huggingface.co/Helsinki-NLP/opus-mt-ru-en}{\texttt{opus-mt-ru-en}} for Russian, and \href{https://huggingface.co/Helsinki-NLP/opus-mt-tc-big-fr-en}{\texttt{opus-mt-tc-big-fr-en}} for French.}
    \item \textbf{English-side QA-SRL parsing:} The translated English sentence is passed through the QA-SRL parser to generate English question-answer pairs anchored to verbal and nominal predicates.
    \item \textbf{Word alignment:} Using  \emph{word alignment} (\S\ref{subsec:word_alignment}) between the English translation and the original target-language sentence, we map English predicates to their target-language counterparts and project English answer spans onto $\mathcal{L}$.
    \item \textbf{Constrained back-translation:} Generated English questions are translated back into~$\mathcal{L}$ using a \emph{predicate-preserving constrained translation} procedure (\S\ref{subsec:constrained_translation}), ensuring the aligned predicate appears verbatim in the question.
\end{enumerate}

An illustrative complete example of the projection pipeline can be found in Appendix \ref{app:projection_example}.
This process yields parallel QA-SRL annotations (sentence, predicate, question, answer) for any language with a basic linguistic resources and a translation system, enabling the creation of large-scale training sets for multilingual QA-SRL parsers.

\subsection{Word Alignment}\label{subsec:word_alignment}


Word alignment links predicates and arguments between the English translation and the target-language sentence. 
We employ SimAlign \citep{jalili-sabet-etal-2020-simalign}, an unsupervised mBERT-based method that frames alignment as bipartite graph matching. 
The alignment is used at two key modules:
\begin{enumerate}[noitemsep, topsep=0pt]
    \item \textbf{Predicate alignment:} For each English predicate captured by the QA-SRL parser, we locate its aligned token(s) in the target-language sentence. This keeps the scope of captured predicates roughly equivalent to the English parser's scope, covering most verbs and deverbal nominalizations in $\mathcal{L}$ (see Appendix \ref{app:predicate_detection} for full details regarding predicate identification).
    \item \textbf{Answer span projection:} Each English answer span is mapped to the original sentence in $\mathcal{L}$ by locating the aligned tokens and extracting the smallest continuous span that contains them (see Appendix~\ref{appendix:span_heuristic} for further details about the answer projection heuristics).
\end{enumerate}

\subsection{Predicate Preserving Constrained Translation}
\label{subsec:constrained_translation}

After generating English QA pairs with the QA-SRL parser and aligning the target-language predicate, we translate the questions into~$\mathcal{L}$ while enforcing that the translated question explicitly contains the aligned predicate, preserving the QA-SRL format. 
We implement this using language-specific LLMs prompted with few-shot examples instructing the model to produce fluent translations of English questions confined to particular $\mathcal{L}$ predicates. 
This procedure is applied uniformly to verbal (QA-SRL) and nominal (QANom) predicates. 
An illustrative example is provided in Appendix~\ref{appendix:constrained_translation}.

The resulting projected annotations form the basis for the multilingual datasets described in Section~\ref{sec:dataset_creation}, which in turn are used to fine-tune and evaluate target-language QA-SRL parsers presented in Section~5.


\begin{table*}[t]
\centering
\renewcommand{\arraystretch}{1.4}
\small  
\begin{tabular}{|l|c|c|c|c|c|c|c|c|c|}
\hline
& \multicolumn{3}{c|}{\textbf{Train}} & \multicolumn{3}{c|}{\textbf{Dev}} & \multicolumn{3}{c|}{\textbf{Test}} \\
\textbf{Language} & \textbf{Sents} & \textbf{Pred.} & \textbf{QAs} & \textbf{Sents} & \textbf{Pred.} & \textbf{QAs} & \textbf{Sents} & \textbf{Pred.} & \textbf{QAs} \\
\hline
\textbf{Hebrew}  & 13,233 & 33,275 & 80,210  & 64  & 146 & 368  & 116 & 342 & 793 \\
\textbf{Russian} & 18,804 & 42,518 & 102,197 & 98  & 258 & 689  & 228 & 603 & 1,582 \\
\textbf{French}  & 18,181 & 38,157 & 97,336  & 51  & 124 & 259  & 144 & 288 & 616 \\
\hline
\end{tabular}
\caption{Number of sentences, predicates, and QA pairs in the projected QA-SRL datasets for Hebrew, Russian, and French.
}
\label{NewDataStat}
\end{table*}

\section{Dataset Creation}
\label{sec:dataset_creation}

To train and evaluate multilingual QA-SRL models, we construct automatically projected QA-SRL datasets for three languages and create a manually curated gold-standard evaluation set.  
This section describes the target languages, source corpora, and annotation pipeline used to build these resources.

\subsection{Languages and Corpora}
\label{subsec:languages_corpora}

To evaluate our projection pipeline across diverse linguistic settings, we target three typologically distinct languages: Hebrew, Russian, and French.
This selection stresses the method under varying structural and resource conditions --- Hebrew as a medium-to-low resource, Semitic language with rich morphology and a revived modern profile; Russian with its flexible word order and extensive case system; and French as a resource-rich Romance language providing structural contrast.

For each language, we extract sentences from   multiple Universal Dependencies (UD) corpora \citep{nivre-etal-2016-universal-v1,de-marneffe-etal-2021-universal}, which provide tokenization and part-of-speech tags essential for predicate identification (see Appendix \ref{app:predicate_detection}).  
Specifically, we use \textit{HTB}, \textit{IAHLTwiki}, and \textit{IAHLTknesset} for Hebrew; \textit{SynTagRus}, \textit{GSD}, \textit{Taiga}, and \textit{PUD} for Russian; and \textit{GSD}, \textit{Sequoia}, \textit{ParisStories}, and \textit{PUD} for French.  
These corpora span diverse genres including news, Wikipedia, government proceedings, blogs, and spoken narratives, providing a broad linguistic sample for each language.

We apply our QA-SRL projection pipeline on the compiled corpora to attain linguistic specific QA-SRL annotations. 
Table ~\ref{NewDataStat} presents descriptive statistics of our projected datasets.

\subsection{Evaluation set annotation}

To evaluate projected QA-SRL annotations, we created a gold-standard dataset through manual human annotation. This served two purposes: (1) providing a reliable benchmark for QA-SRL models in the target languages, and (2) assessing the accuracy of our cross-lingual projection pipeline.

We sampled projected QA pairs from multiple corpora in each language and reviewed them via a multi-step manual process. Annotators corrected substantive errors (e.g., incorrect or missing questions/answers), removed invalid pairs caused by translation or alignment failures, and added missing but valid QAs. Minor stylistic edits were applied for clarity, while all substantial modifications (e.g., replacing or adding QA pairs) were explicitly flagged for later evaluation of the projection algorithm.

Hebrew and French datasets were annotated by the authors, while Russian annotation was carried out by a native speaker experienced in QA-SRL tasks, compensated at \$15/hour. The average annotation rate was ~90 QA pairs per hour. All annotators ensured grammaticality, faithfulness to the source sentence, and full coverage of the predicate’s semantic arguments.

\subsection{Quality Assessment of Data}

\subsubsection{Evaluation Criteria}
We evaluate QA-SRL annotations using standard SRL metrics adapted to the QA-based format, focusing on two subtasks: \textbf{Unlabeled Argument Detection} and \textbf{Labeled Argument Detection}, corresponding to \textit{Argument Detection} and \textit{Label Assignment} in traditional SRL. 
These metrics have been previously adopted in QA-SRL and QANom evaluations \citep{roit2020qasrl-gs, klein-etal-2020-qanom} and are instantiated in our automatic setup (\S\ref{subsec:auto-eval}) as Argument Match and Question Match, respectively.

Unlabeled Argument Detection checks whether the predicted answer span covers a valid argument regardless of question phrasing, while Labeled Argument Detection additionally requires matching the role assignment conveyed by the gold question.

\subsubsection{Evaluating the Training Set}

Table \ref{tab:auto_train_eval} reports Precision, Recall, and F1 for Hebrew, Russian, and French. High Unlabeled F1 across all languages shows that the projection pipeline captures most predicate arguments, even in morphologically rich, structurally diverse settings. Since we focus on generating full QA sets for each predicate, this demonstrates effective reconstruction of core argument structures.

Recall remains high in both settings, indicating broad coverage of arguments and questions --- critical for training, where missing information is more harmful than occasional noise. As expected, Labeled scores are lower due to stricter matching criteria but remain strong enough to support reliable supervision.

By combining UD corpora, MT, word alignment, and constrained question translation, we built high-quality Hebrew, Russian, and French training data. Manual evaluation confirmed robust predicate-argument coverage, forming a solid foundation for developing multilingual QA-SRL parsers.

\begin{table}[t!]
\centering
\large
\renewcommand{\arraystretch}{1.1}
\resizebox{\columnwidth}{!}{%
\begin{tabular}{|l|ccc|ccc|}
\hline
& \multicolumn{3}{c|}{\textbf{Unlabeled AD}} 
& \multicolumn{3}{c|}{\textbf{Labeled AD}} \\
\hline
\textbf{Language} 
& \textbf{P} & \textbf{R} & \textbf{F1} 
& \textbf{P} & \textbf{R} & \textbf{F1} \\
\hline
\textbf{Hebrew} 
& 67.8 & 94.7 & 79.0 
& 57.4 & 93.8 & 71.2 \\
\textbf{Russian} 
& 70.3 & 87.9 & 78.1 
& 47.3 & 83.3 & 60.3 \\
\textbf{French} 
& 81.3 & 93.5 & 87.0 
& 60.8 & 91.6 & 73.1 \\
\hline
\end{tabular}%
}
\caption{Manual evaluation results of the projected training sets.}
\label{tab:auto_train_eval}
\end{table}

\paragraph{Error sources in the projection pipeline.}
To better understand the quality of the projected training data, we conducted a targeted analysis of error sources across the projection pipeline.
Overall, we find that most errors originate from the English QA-SRL/QANom parser itself, which serves as an effective upper bound on projection quality and is known to be recall-limited \citep{roit2020qasrl-gs,klein-etal-2022-qasem}.
In contrast, errors introduced during back-projection are comparatively rare.
Predicate alignment errors occur infrequently (4 misaligned predicates out of 161 examined in the manually curated gold set), and predicate-preserving constrained translation achieves a high success rate, effectively mitigating predicate drift.
Remaining back-projection errors primarily involve minor answer-span boundary inaccuracies due to alignment gaps, rather than systematic semantic distortions.
Taken together, this analysis indicates that the projected datasets preserve predicate--argument structure reliably, and that further gains in projection quality will depend primarily on advances in English QA-SRL parsing rather than on the cross-lingual transfer mechanisms themselves.

\section{Models and Evaluation}
\label{sec_model_evaluation}

In this section, we evaluate the quality of our QA-based semantic parsers across three target languages. We first describe the automatic evaluation framework (\S\ref{subsec:auto-eval}), followed by the experimental setup and models used (\S\ref{subsec:exp-setup}), and conclude with the main results and analysis (\S\ref{subsec:results}).

\subsection{Automatic Evaluation}
\label{subsec:auto-eval}



To enable large-scale evaluation of model predictions, we significantly adapt the standard protocol from prior QA-SRL work \citep{roit2020qasrl-gs, klein-etal-2020-qanom}. We operationalize the two subtasks defined in Section~\ref{sec:dataset_creation} via a two-stage automatic procedure:   
\textbf{Argument Match}, aligning predicted and gold answer spans, and \textbf{Question Match}, assessing whether the predicted question expresses the same semantic role as the reference.



For Argument Match, we adopt the token-level alignment method of \citet{roit2020qasrl-gs}. 
Predicted and gold answer spans are connected in a bipartite graph weighted by token-level \textit{Intersection-over-Union (IOU)}; edges below a calibrated threshold $\tau = 0.5$ are discarded.
Then, a maximal bipartite matching algorithm selects the best one-to-one alignments between predicted and gold spans.  
Spans that align above the threshold are treated as matches, while the rest are considered mismatches.
Threshold tuning and further details are elaborated in Appendix~\ref{appendix:iou-threshold}.

In the Question Match stage, we impose a stricter QA-to-QA match criterion and evaluate the questions associated with matched answer spans, which implicitly encode the argument's semantic role.  
Prior QA-SRL work has struggled to establish reliable question-level metrics, with existing approaches tied to English-specific templates and ill-suited for multilingual, free-form questions.

To address this, we introduce two complementary language-agnostic \textbf{Question Match} criteria:  
(1) \textbf{Exact Match}, requiring string identity with the gold question, and  
(2) \textbf{Semantic Match}, which scores a question as correct if its embedding is semantically equivalent to the gold according to a SentenceTransformers paraphrase model (cosine similarity~$\geq 0.78$).  
Appendix~\ref{appendix:semantic-threshold} details the model and threshold calibration.

This two-stage evaluation combines strict and relaxed signals for assessing predicate-argument structure, providing a scalable approximation to human judgment across languages.

\subsection{Experimental Setup and Models}\label{subsec:exp-setup}



To assess the effects of scale, supervision, and language specialization, we evaluate three model families:
(i) \textbf{instruction-tuned LLMs} (hundreds of billions of parameters) serving as few-shot no-projection baselines;
(ii) a \textbf{multilingual 8B language model}, evaluated both in its in-context learning (ICL) mode and after LoRA fine-tuning (FT); and
(iii) \textbf{language-specific 7B models}, likewise evaluated before and after fine-tuning.

This design enables us to isolate the effects of fine-tuning and to compare multilingual versus monolingual pre-training at a similar scale.

We employ \textbf{GPT-4o} \citep{openai2024gpt4o} and \textbf{LLaMA-4-Maverick}  \citep{meta2025llama4} as few-shot instruction-tuned baselines, prompted with two language-specific examples for each predicate type (full prompts are provided in Appendix \ref{app:icl_prompt}). These LLMs, while requiring substantially more inference-time computation, provide a strong reference point without any task- or language-specific adaptation.

At the 8B scale, we employ \textbf{LLaMA-3-8B} \citep{llama3modelcard} as our multilingual backbone and compare it directly with monolingual language models of similar size.
For both types, we evaluate (a) their in-context performance and (b) their LoRA-adapted variants, allowing us to examine how fine-tuning on projected data affects performance under a consistent architecture and parameter budget.

For the language-specific models, we use \textbf{DictaLM} for Hebrew \citep{shmidman2024adaptingllmshebrewunveiling}, \textbf{SambaLingo} for Russian \citep{csaki2024sambalingo}, and \textbf{Claire} for French \citep{louradour2024claire}.
Their single-language pretraining provides specialized syntactic and lexical priors that we hypothesize to enhance QA-based semantic parsing. 
All models are LoRA finetuned on the our automatically projected datasets (\S \ref{sec:dataset_creation}), with hyperparameters selected on held-out development sets (full details in Appendix~\ref{appendix:finetuning}).

To illustrate the task and expected outputs, Table~\ref{tab:qa-examples} presents model-generated QA pairs for the sentence ``The committee \textbf{approved} the new policy'', translated into Hebrew, Russian and French.
We next present the evaluation results, comparing performance across languages and model types.






\subsection{Results}
\label{subsec:results}


\begin{table*}[h!]
\centering
\renewcommand{\arraystretch}{1}
\begin{tabular}{@{}llccc@{}}
\toprule
\textbf{Model} & \textbf{Metric} & \textbf{Hebrew} & \textbf{Russian} & \textbf{French} \\
\midrule
\multirow{3}{*}{GPT-4o (ICL)} 
    & Unlabeled Match & \textbf{66.3} & \textbf{67.9} & 63.2 \\
    & Exact Match     & 10.5 & 14.1 & 10.2 \\
    & Semantic Match  & 38.3 & 47.0 & 41.8 \\
\midrule
\multirow{3}{*}{LLaMA-4-Maverick (ICL)} 
    & Unlabeled Match & 61.6 & 60.1 & 52.6 \\
    & Exact Match     & 12.0 & 12.8 & 7.5 \\
    & Semantic Match  & 38.8 & 45.0 & 30.7 \\
\midrule\midrule
\multirow{3}{*}{LLaMA-3-8B (ICL)} 
    & Unlabeled Match & 43.3 & 51.8 & 47.8 \\
    & Exact Match     & 2.0  & 4.5  & 3.5 \\
    & Semantic Match  & 20.4 & 36.5 & 26.7 \\
\midrule
\multirow{3}{*}{LLaMA-3-8B (FT)} 
    & Unlabeled Match & 56.5 & 61.7 & 60.5 \\
    & Exact Match     & 21.2 & 19.2 & 19.1 \\
    & Semantic Match  & 42.7 & 53.1 & 42.6 \\
\midrule\midrule
\multirow{3}{*}{Language-Specific (ICL)} 
    & Unlabeled Match & 51.6 & 42.3 & 32.5 \\
    & Exact Match     & 8.1  & 4.5  & 2.9 \\
    & Semantic Match  & 31.9 & 31.5 & 16.4 \\
\midrule
\multirow{3}{*}{Language-Specific (FT)} 
    & Unlabeled Match & 63.0 & 65.7 & \textbf{65.9} \\
    & Exact Match     & \textbf{30.1} & \textbf{25.4} & \textbf{23.1} \\
    & Semantic Match  & \textbf{51.7} & \textbf{58.1} & \textbf{57.2} \\
\bottomrule
\end{tabular}
\caption{F1 scores of baseline and fine-tuned models across Hebrew, Russian, and French.
Double horizontal rules delineate three blocks: 
(1) The upper block lists strong large-scale baseline systems (GPT-4o and LLaMA-4-Maverick). 
(2) The middle block reports a mid-scale multilingual model before and after fine-tuning on our projected dataset. 
(3) The lower block presents language-specific models that are comparable in scale to the middle block, comparing their in-context and fine-tuned variants per language.}
\label{tab:LLM_F1_Results}
\end{table*}

Table~\ref{tab:LLM_F1_Results} reports performance across Hebrew, Russian, and French for all model types, evaluated with Unlabeled Match, Exact Match, and Semantic Match.\footnote{For reference, the English QA-SRL parser used to generate the projected annotations reports 75.9~F1 on QA-SRL and 72.4~F1 on QANom under unlabeled argument detection \citep{cattan2024localizing}. These values provide a loose upper bound on parsers trained over annotations projected from predicted English QA-SRL/QANom.}

Large instruction-tuned models, particularly GPT-4o, achieve strong scores on Unlabeled Match in all languages, confirming their ability to detect salient argument spans without task-specific supervision.
In Hebrew and Russian, their performance approaches or slightly exceeds that of fine-tuned models, though the gap is small.
In French, however, the language-specific fine-tuned model surpasses all baselines, showing that targeted supervision can still improve even surface-level argument detection with a much smaller model.

For question-based metrics, the contrast is sharper: fine-tuned models, especially language-specific ones, consistently outperform few-shot prompting by a wide margin on both Exact and Semantic Match.
This pattern holds across all languages, underscoring that while large instruction-tuned models can approximate argument boundaries, generating semantically appropriate questions benefits substantially from explicit supervision.
Given the crucial role of question precision in conveying the argument’s semantic role, these results highlight the limitations of in-context learning and reinforce the importance of our training-data projection pipeline in enabling cross-linguistic supervision.
We note that Semantic Match is a deliberately conservative labeled metric, calibrated for high precision and therefore prone to false negatives (Appendix~\ref{appendix:semantic-threshold}); as a result, labeled argument performance is systematically underestimated by this metric.

Among fine-tuned systems, language-specific models perform best overall.
Their consistent advantage over the multilingual LLaMA-3-8B variant underscores the value of language specialization, particularly for producing precise, semantically-oriented annotations.
Overall, these findings highlight that language-aligned models are key to achieving robust QA-based semantic parsing across diverse languages.

\paragraph{Statistical significance.}
To assess whether the relatively small manually curated test sets affect the robustness of the observed trends, we conducted 10{,}000-iteration paired bootstrap significance tests by resampling predicates and recomputing F1 scores from their true positive, false positive, and false negative contributions (approximately 300–550 predicates per language).
For Semantic Match F1, the advantage of the language-specific fine-tuned models over GPT-4o in-context learning is statistically significant in all three languages ($p<0.001$).
In contrast, for Unlabeled Match, differences between systems are not statistically significant in any language ($p=0.08$--$0.17$), supporting the conclusion that the compared approaches yield broadly similar argument detection performance while diverging substantially in question quality.

\paragraph{Sensitivity to the Semantic Match threshold.}
To assess the sensitivity of our comparative results to the similarity threshold used by the Semantic Match metric, we re-evaluated all models across all languages using thresholds ranging from 0.70 to 0.90.
As expected, absolute Semantic Match scores decrease gradually as the threshold becomes stricter.
Crucially, however, the relative performance pattern remains stable throughout this range: the performance gap between fine-tuned models and GPT-4o is essentially unchanged for all three languages.
This analysis indicates that while the automatic metric exhibits moderate sensitivity to threshold choice in absolute terms, our comparative conclusions regarding model performance are highly robust to this hyperparameter.

\paragraph{Manual error analysis of question generation.}
To better characterize labeled evaluation errors, we manually analyzed cases in which predicted answers passed the IOU-based Argument Match stage but failed Semantic Match.
Across 50 such instances from the Hebrew test set, approximately half were semantically acceptable, corresponding to valid paraphrases or legitimate alternate questions induced by minor span differences.
The remaining cases reflect genuine modeling errors, primarily involving incorrect predicate realization or role targeting.
Overall, this analysis indicates that Semantic Match failures often reflect conservative matching rather than spurious predictions, leading to systematic underestimation of labeled performance.
Detailed taxonomy and examples appear in Appendix~\ref{appendix:error_analysis}.

\paragraph{} 
\paragraph{} 
In sum, few-shot prompting with large general-purpose models remains competitive for identifying argument spans, but consistently falls short in generating semantically faithful questions.
Across languages, evaluation metrics, significance testing, and threshold sensitivity analyses all converge on the same conclusion: fine-tuned models trained on language-specific projected data deliver substantially higher-quality semantic role representations.
Notably, these gains are achieved by models that are orders of magnitude smaller than the instruction-tuned LLM baselines, reinforcing our central claim that targeted supervision and language adaptation remain essential for accurate, interpretable, and efficient QA-based semantic parsing in multilingual settings.

\section{Conclusion}
\label{sec_conclusion}

We introduced a scalable projection approach for producing QA-based predicate-argument annotations in new languages.
By reusing an English QA-SRL parser within a pipeline of constrained translation and  word alignment, our method generates high-quality training data for structurally diverse languages --- demonstrated on Hebrew, Russian, and French.
Fine-tuned models trained on this projected supervision achieve strong, language-specific predicate-argument parsing performance, approaching English accuracy on unlabeled argument detection and substantially outperforming few-shot prompted LLMs in question generation accuracy.
These findings underscore the enduring value of explicit, task-specific supervision even in the era of powerful general-purpose models.

Beyond empirical results, our contribution lies in making QA-based predicate-argument analysis broadly attainable across languages with minimal human annotation.
By turning QA-SRL into a practical projection vehicle, this work extends the reach of interpretable, semi-structured predicate-argument representations to low- and mid-resource languages --- laying the foundation for multilingual applications such as fine-grained semantic analysis and content-based generation evaluation.

\section{Limitations}\label{sec_limitation}

Our core methodological contribution --- the QA-SRL projection pipeline --- relies on a sequence of model-dependent components, including translation, English QA-SRL parsing, word alignment, and constrained question translation.  
While this modular design enables scalable annotation transfer, it also introduces susceptibility to error propagation: inaccuracies at any stage (e.g., translation drift, parser errors, or alignment mismatches) can cascade and compound in the final projected annotations.  
As a result, the quality of the generated datasets and trained parsers is tightly linked to the robustness of each component, which may pose challenges when extending the approach to diverse languages or noisy real-world data.

While our experiments focus on Hebrew, Russian, and French, the method generalizes to any language with basic POS tagging, translation and alignment tools into English, and a pretrained language model capable of basic in-context tasks. Languages lacking these resources may pose quality challenges that are not addressed by the current proposed method. Thus, highly polysynthetic or low-resource languages remain an open direction for future work.



\section*{Acknowledgements}\label{sec_acknowledgements}
We would like to thank our reviewers for their
constructive suggestions and insightful comments.
This research has been funded by a grant from the Planning and Budgeting Committee (PBC) of the Israeli Council of Higher Education (CHE), and by a grant from the Israeli Science Foundation (ISF grant no.\ 670/23), for which we are grateful.


\bibliography{custom}

\appendix
\section{English QA-SRL/QANom Parser}
\label{app:QASem-parser}

For generating English annotations, we use the joint QA-SRL/QANom parser introduced by \citet{klein-etal-2022-qasem}, which models both tasks within a unified text-to-text formulation.
The model is trained jointly on verbal and nominal predicates, but at inference time operates on a single marked predicate per sentence, generating natural language questions and corresponding answer spans for that predicate.
This design leverages the natural alignment between QA-based semantics and sequence-to-sequence modeling while allowing consistent treatment of both QA-SRL and QANom.

We adopt the state-of-the-art implementation first released by \citet{roit-etal-2024-explicating} and later improved by \citet{cattan2024localizing}, which uses a larger T5 variant (T5-XL 3B) trained on the combined QA-SRL \citep{fitz2018qasrl} and QANom \citep{klein-etal-2020-qanom} corpora.

This parser\footnote{Available at \url{https://github.com/plroit/qasem_parser}.} achieves state-of-the-art labeled and unlabeled argument detection on QA-SRL and QANom benchmarks and provides the English predicate-argument structures projected into Hebrew, Russian, and French in our cross-lingual pipeline.
\section{Illustrative Example: Limitations of Direct Surface Translation}
\label{app:surface_translation_limitations}

A straightforward yet naive approach to obtaining QA-SRL annotations in a new target language is a direct round-trip translation method: the target sentence is translated into English, existing QA-SRL tools are applied to produce English question-answer pairs, and these QAs are then directly translated back into the target language.
To demonstrate why this method is insufficient for cross-lingual QA-SRL projection, consider the following Hebrew sentence and its translation.

\begin{quote}
\textbf{Source Sentence (Hebrew):} \\
{\scshape{\scriptsize{\hebrew{לירושלים. סעו ישראל: ילדי לכל אומר אני ולכן}}}}

\textbf{English Translation :} \\
And so I say to all the children of Israel: Go to Jerusalem.

\textbf{English QA} (QA-SRL parser on \textit{English Translation}): \\
Q: \emph{Who should go somewhere?} \\
A: \quad `all the children of Israel' 

\textbf{Surface Hebrew Translation of \textit{English QA}}: \\
Q: {\scshape{\scriptsize{\hebrew{לאנשהו? ללכת צריך מי}}}} \\
A:{\quad \scshape{\scriptsize{\hebrew{ישראל' בני 'כל}}}} \end{quote}

\noindent
\textbf{Issues observed:}
\begin{itemize}
    \item \textbf{Predicate drift:} The Hebrew predicate \emph{``סעו''} (``go/plural imperative'' --- in the sense of a drive) from the original sentence is missing from Hebrew translation of the English QA, replaced by a different root \emph{``ללכת''} (``to walk''). As a result, the semantics of the question is misaligned with the original context of the source sentence.
    \item \textbf{Argument mismatch:} The original Hebrew span {\scshape{\scriptsize{\hebrew{ישראל' ילדי 'כל}}}} (``all the children of Israel'') is altered to {\scshape{\scriptsize{\hebrew{ישראל' בני 'כל}}}} (``all the sons of Israel''), which is semantically different from the original argument. 
    \item \textbf{Distorted QA structure:} In addition to the semantic drift of the back-translated QA, it also no longer corresponds precisely to the original Hebrew sentence surface form. QA-SRL is designed to provide a semi-structured  representation of predicate--argument relations through the correspondence of answers to sentence spans (arguments) and the alignment of the question's predicate with its occurrence in the sentence. Breaking these alignments hinders the downstream use of QA-SRL as a semantic representation or decomposition of target language sentences.       
\end{itemize}

\noindent
To conclude, this example highlights that naive direct QA translation introduces both \textbf{predicate replacements} and \textbf{argument span divergences}. Therefore, relying solely on English tools via machine translation cannot ensure faithful QA-SRL in new languages. Our approach instead incorporates constrained translation and word alignment to preserve predicate alignment and maintain accurate span mappings between the sentence and the QAs.

\section{Illustrative Example of the Projection Pipeline (French)}
\label{app:projection_example}

This appendix presents a step-by-step example of our multilingual QA-SRL projection pipeline for a French sentence.

\paragraph{Source sentence (French):}
\textit{Je me suis finalement abstenue en ce qui concerne le vote pour un certain nombre de raisons.}

\paragraph{Translation to English:}
\textit{Finally, I abstained from voting for a number of reasons.}

\paragraph{Step 1: Predicate identification (English)}
The English QA-SRL parser detects the predicate \textbf{``abstained''} (index = 3) as a verbal predicate.

\paragraph{Step 2: Predicate alignment}
Using \textit{word alignment}, the corresponding French predicate is identified as \textbf{``abstenue''}.

\paragraph{Step 3: English QA-SRL output}
The parser produces the following English question–answer pairs:

\small
\setlength{\tabcolsep}{4pt}  
\renewcommand{\arraystretch}{1.1}

\noindent\begin{tabular}{p{0.64\linewidth} p{0.30\linewidth}}
\toprule
\textbf{Question} & \textbf{Answer} \\
\midrule
Who abstained from something? & I \\
What did someone abstain from? & voting \\
Why did someone abstain from something? & for a number of reasons \\
\bottomrule
\end{tabular}
\normalsize

\paragraph{Step 4: Question translation}
Each question is translated into French using the \textit{ constraint translation}, ensuring that the predicate is faithfully preserved in the target language.

\paragraph{Step 5: Answer span alignment}
Using \textit{word alignment} between English and French sentences, we identify the corresponding spans in the French text for each English answer.
The aligned spans are then paired with the translated French questions, yielding the projected QA-SRL annotations for the target language:

\small
\setlength{\tabcolsep}{4pt}
\renewcommand{\arraystretch}{1.1}

\noindent\begin{tabular}{p{0.64\linewidth} p{0.30\linewidth}}
\toprule
\textbf{Question (FR)} & \textbf{Answer} \\
\midrule
Qui s’est abstenu de quelque chose ? & Je \\
De quoi quelqu’un s’est-il abstenu ? & du vote \\
Pourquoi quelqu’un s’est-il abstenu de quelque chose ? & pour un certain nombre de raisons \\
\bottomrule
\end{tabular}
\normalsize

\paragraph{Summary}
This example illustrates the three core operations of the projection pipeline:
\begin{itemize}
    \item \textbf{Predicate alignment} between English and target-language verbs.
    \item \textbf{Constrained question translation} that preserves the original predicate in the question during translation.
    \item \textbf{Answer span alignment} ensuring that argument spans in the target-language sentence correspond accurately to the English span answer.
\end{itemize}
Together, these steps produce high-quality projected supervision for fine-tuning target-language QA-SRL parsers.

\section{Identifying Predicates in Target Language}
\label{app:predicate_detection}

Our projection algorithm transfers QA pairs generated on the English translation back to the original sentence in the target language~$\mathcal{L}$. 
The QA-SRL and QANom parsers we train assume a pre-specified predicate, which must be either a verb or a deverbal nominalization. 
Section~\ref{sec_model_evaluation} reports QA generation performance under this setting with gold predicates provided.
Because predicate detection is a prerequisite for training and inference yet introduces distinct challenges --- especially for eventive nominalizations --- this appendix details our strategy for identifying predicates in both the projected training data and the final parsers.

\paragraph{Challenges.} 
Identifying predicates in the target language is non-trivial. 
While verbal predicates can be reliably detected via POS tagging, distinguishing deverbal eventive nominalizations from non-predicative nouns is substantially harder \citep{klein-etal-2020-qanom}. 
This motivates our reliance on Universal Dependencies (UD) corpora, which supply both tokenization (necessary for alignment) and POS tags.

\paragraph{Predicate selection in training-set projection.} 
For the projected training data, we rely on English-side predicates produced by the QA-SRL parser and their aligned tokens in~$\mathcal{L}$. 
Since lexical category often shifts in translation (e.g., a verb in one language aligning to a nominalization in another), we take the union of QA pairs generated for both English verbs and nominalizations. 
Because the English parser incorporates a trained nominalization detector \citep{klein-etal-2020-qanom}, this union covers most verbal and nominal predicates in the target language.  
We filter out English predicates whose aligned tokens in~$\mathcal{L}$ are neither verbs nor nouns; for nouns we further require the aligned token to be identified as a deverbal nominalization by our classifier (described below).

\paragraph{Predicate identification at inference time.} 
For evaluation against the manually corrected gold standard, we focus purely on QA generation and rely on gold predicates identified during the training-set projection stage using alignment and the nominalization classifier.  
For the released parsers, and for new sentences lacking UD annotations, we:  
(i) apply SpaCy for tokenization and POS tagging to detect verbs and nouns;  
(ii) use our nominalization classifier over candidate nouns to separate static entities from deverbal predicates;  
(iii) pass the identified predicates to the QA-SRL or QANom parser for question-answer generation.

\paragraph{Nominalization classification via in-context learning.} 
To separate eventive nominalizations from non-predicative nouns, we employ a large language model in a few-shot in-context learning (ICL) setup. 
The prompt provides examples of both static entity nouns and deverbal predicates, for instance (examples shown in French):


\begin{itemize}[noitemsep, topsep=0pt]
    \item assiette: nom commun 
    \item invitation: nom d'action 
    \item comité: nom commun 
    \item permission: nom d'action 
    \item libération: nom d'action 
\end{itemize}

Given a new noun, the model predicts whether it functions as an eventive nominalization or a static entity. 
This approach is effective in most cases but has inherent limitations: it does not fully capture context-sensitive uses. 
For example, the Hebrew noun{\scshape{\scriptsize{\hebrew{{"ארגון"}}}}} (“organization”) can be a static entity in {\scshape{\scriptsize{\hebrew{העולמי הבריאות ארגון}}}} (“the World Health Organization”), but an eventive nominalization in {\scshape{\scriptsize{\hebrew{בחדר החפצים ארגון}}}} (“the organizing of the room’s items”).
Context-aware extensions of this classifier are left to future work.


\section{Answer Span Postprocessing Heuristics}
\label{appendix:span_heuristic}

To improve answer span quality, we applied three lightweight heuristics during post-processing:

First, if the predicted span was noncontiguous (e.g., due to alignment gaps), we expanded it to the minimal contiguous span covering all tokens.

\textbf{Example:}  
For the Hebrew (tokenized) sentence:
{\scshape{\scriptsize{\hebrew{אחוזים שני של הפרש ב השבוע נוצח הוא}}}}
(“He was defeated this week by a margin of two percent”),  
given the question {\scshape{\scriptsize{\hebrew{נוצח? מישהו איך}}}} (How was someone defeated?),  
The align answer span was {\scshape{\scriptsize{\hebrew{אחוזים" שני של "ב}}}} (by a of two percent), missing the word {\scshape{\scriptsize{\hebrew{``הפרש''}}}} (“margin”).  
Our heuristic filled the alignment gap and produced the corrected answer{\scshape{\scriptsize{\hebrew{אחוזים" שני של "בהפרש}}}} (by a margin of two percent).

Second, for Hebrew only, we iteratively removed function words from the end of the span if they could not plausibly terminate a noun phrase (e.g., prepositions, conjunctions, and definite articles). 
This heuristic was designed to correct alignment artifacts that produced incomplete or ungrammatical span endings. Although the approach applies to other languages (for example, removing 'the' or 'a' in English), we applied it only to Hebrew in this work.

\textbf{Example:}  
For the Hebrew sentence: \\
{\scshape{\scriptsize{\hebrew{2015 באוגוסט פרסם ש מחקר ב ישראל בנק טען כך}}}} 
(“Thus claimed the Bank of Israel in a study published in August 2015.”),  
given the question {\scshape{\scriptsize{\hebrew{פורסם? מה}}}} (What was published?),  
the predicted answer span was \\
{\scshape{\scriptsize{\hebrew{ש" "מחקר}}}} (“a study that”), which ends with the complementizer {\scshape{\scriptsize{\hebrew{``ש''}}}} (“that”), a function word that cannot end a grammatical phrase.  
Our heuristic removed it, yielding the corrected answer {\scshape{\scriptsize{\hebrew{``מחקר''}}}} (“study”).

Third, if the span contained a sentence-internal period or included the predicate token itself, we split the span and retained the longer segment, excluding the problematic element.

These rules improved grammaticality and reduced noise in both training and evaluation.

\section{Predicate Preserving Constrained Translation}
\label{appendix:constrained_translation}

As discussed in Appendix~\ref{app:surface_translation_limitations}, naïve surface translation can lead to predicate drift, where the lexical root of the original verb is replaced during translation.  
In that example, the Hebrew predicate 
{\scshape\scriptsize\hebrew{"סעו"}} (“go” — in the sense of “drive”)  
was replaced by {\scshape\scriptsize\hebrew{"ללכת"}} (“to walk”) in the back-translated question, altering the semantics and breaking the alignment between the question and the original predicate.

To prevent such drift, we apply \textit{predicate-preserving constrained translation}.  
When translating the English QA back into the target language, the LLM receives the English question along with the intended predicate from the source sentence as a lexical constraint.  
The prompt provided to the model is constructed as follows:
\begin{center}
Who should go somewhere? \textbar\ {\scshape\scriptsize\hebrew{לנסוע}}
\end{center}

This explicitly instructs the model to preserve the original predicate root ({\scshape\scriptsize\hebrew{לנסוע}} — “to drive / go”).  
As a result, the model produces the correctly aligned Hebrew question:
{\scshape\scriptsize\hebrew{לאנשהו? לנסוע צריך מי}} \\
(lit. “Who should go somewhere?” — using \textit{go} in the sense of “drive” matching the original Hebrew predicate)

As a result, the predicate remains faithfully preserved in the target-language question.
\section{Argument Matching Procedure and IOU Threshold Calibration}
\label{appendix:iou-threshold}

\paragraph{Argument Matching Procedure.}
To evaluate argument prediction quality, we construct a bipartite graph between predicted and gold answer spans for each predicate instance. 
Each edge is weighted by the token-level Intersection-over-Union (IOU) between the two spans. 
Edges with IOU below a threshold~$\tau$ are discarded to avoid spurious partial matches. 
We then apply a maximal bipartite matching algorithm to select a one-to-one mapping between predicted and gold arguments that maximizes total IOU weight. 
Aligned pairs above~$\tau$ are counted as true positives, while unmatched predicted arguments are treated as false positives and unmatched gold arguments as false negatives. 
This process follows the Unlabeled Argument Detection metric introduced by in \citet{roit2020qasrl-gs} and adopted by subsequent QA-SRL/QANom works \citep{klein-etal-2020-qanom,klein-etal-2022-qasem,roit-etal-2024-explicating,cattan2024localizing}.



\begin{figure}[h]
\centering
\includegraphics[width=\columnwidth, keepaspectratio]{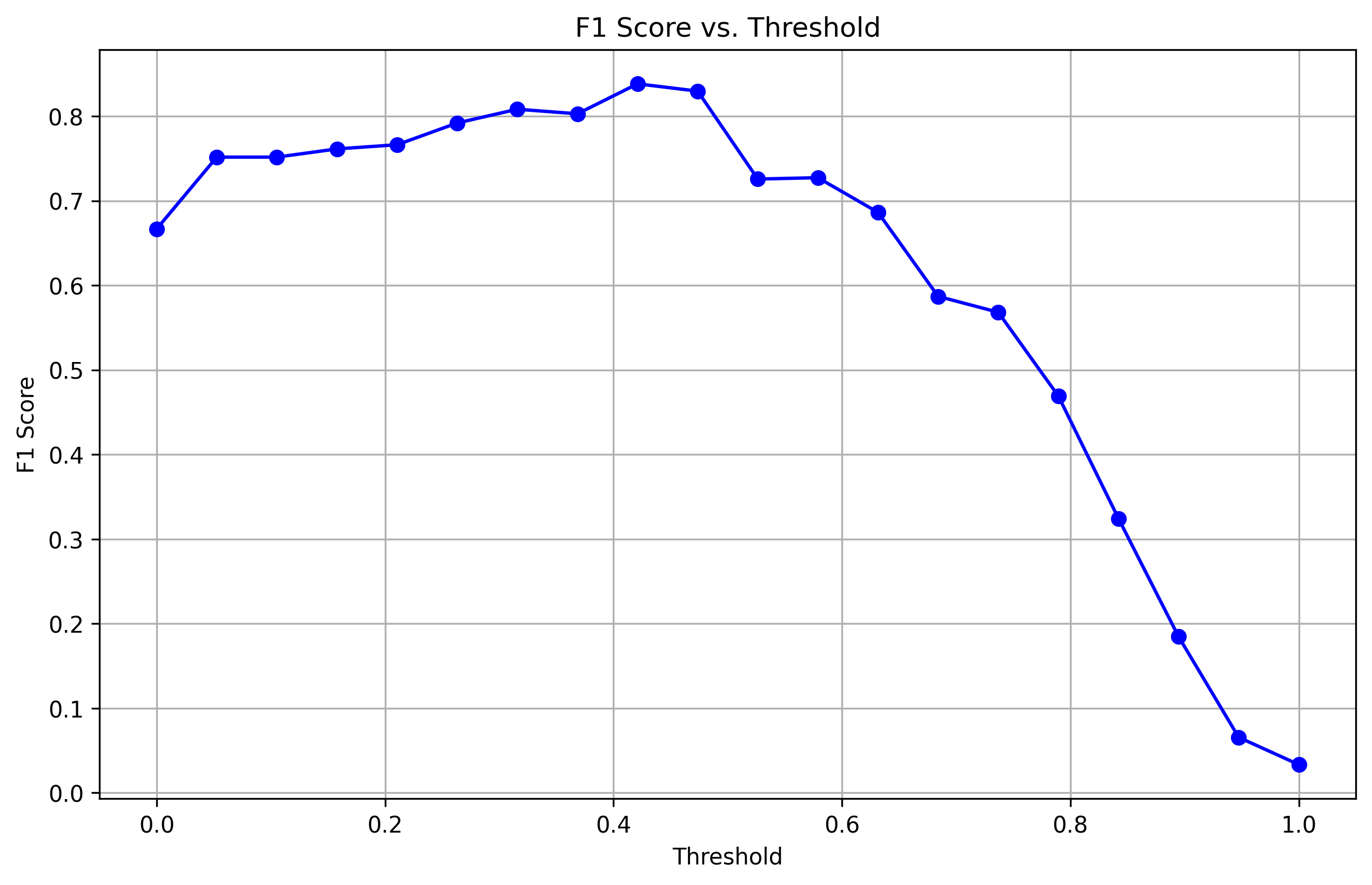}
\caption{F1 score as a function of the IOU threshold}
\label{fig:f1_vs_threshold}
\end{figure}

\begin{figure}[h]
\centering
\includegraphics[width=\columnwidth, keepaspectratio]{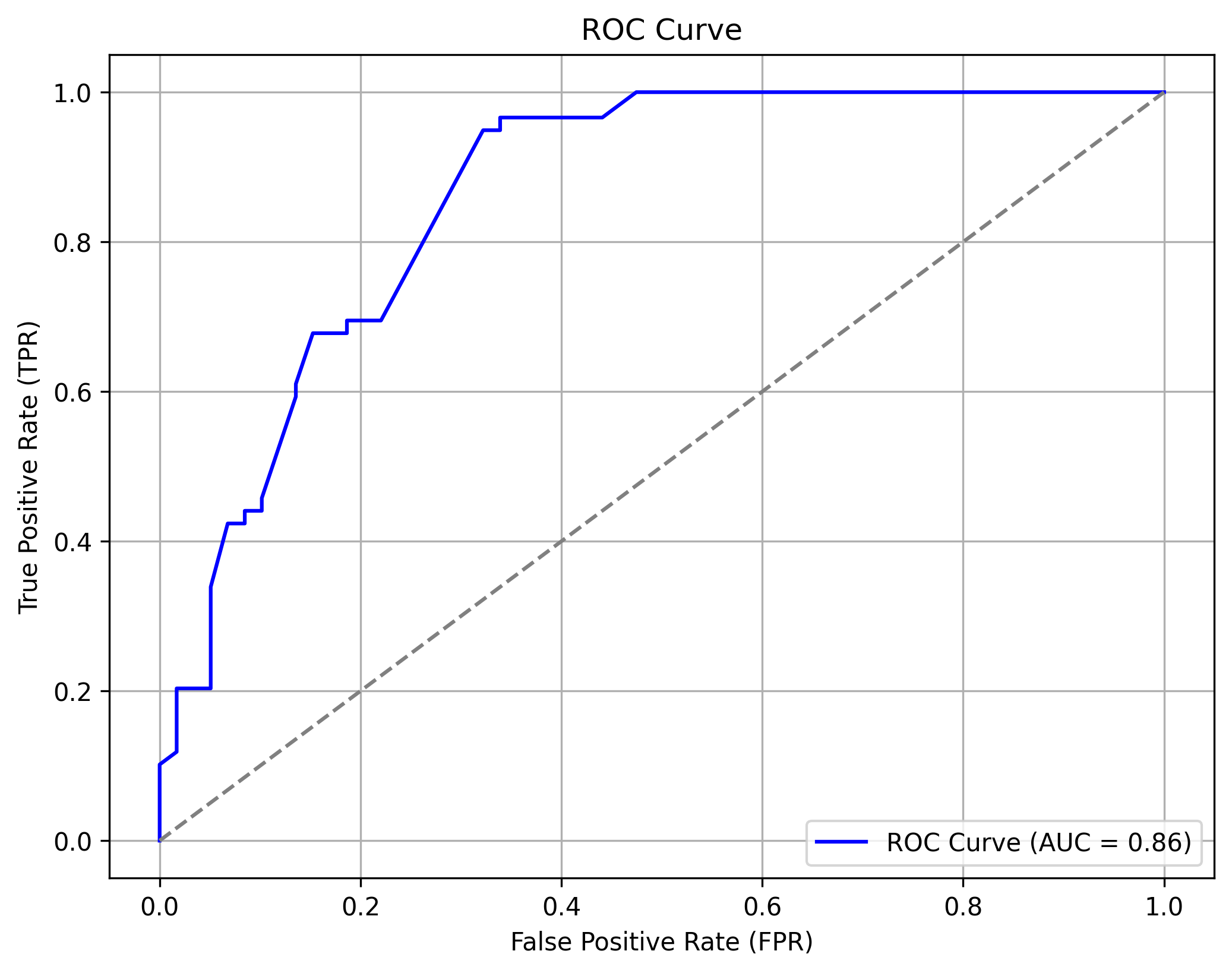}
\caption{ROC curve for span matching decisions}
\label{fig:roc_curve}
\end{figure}

\paragraph{Threshold Calibration.}
Although prior QA-SRL studies \citep{klein-etal-2020-qanom, klein-etal-2022-qasem, roit-etal-2024-explicating, cattan2024localizing} typically use a relatively lenient threshold of $\tau=0.3$, we adopt a stricter value of $\tau=0.5$. 
This was determined via two complementary procedures on a manually annotated validation set: (i) selecting $\tau$ that maximized F1, and (ii) identifying the optimal cutoff from the ROC (Receiver Operating Characteristic) curve of true vs.\ false positive matches. 
Both analyses indicated that $\tau=0.5$ yields the best balance between precision and recall, reducing false positives and improving robustness in morphologically rich and syntactically flexible languages. 
Figures~\ref{fig:f1_vs_threshold} and~\ref{fig:roc_curve} illustrate the calibration results.

\section{Semantic Similarity Model and Threshold Calibration}
\label{appendix:semantic-threshold}

To evaluate labeled predicate--argument correctness in QA-SRL, we require a language-agnostic method for determining whether two questions express the same semantic role.
We use the \textit{paraphrase-multilingual-mpnet-base-v2} model from the \textit{SentenceTransformers} library, which encodes questions into dense vector representations and computes semantic similarity via cosine similarity as a proxy for semantic equivalence.

\paragraph{Threshold calibration.}
To determine an appropriate similarity threshold for this semantic equivalence test, we conducted a calibration study on the Hebrew gold dataset.
For a sample of predicted--gold question pairs, human annotators judged whether the predicted question was semantically equivalent to the reference question.
We evaluated classifier behavior across cosine similarity thresholds ranging from 0.50 to 0.95 (in increments of 0.01), and measured agreement with human judgments.

Rather than optimizing raw accuracy or F$_1$, we selected the threshold that maximized the $F_\beta$ score with $\beta=0.5$, placing greater weight on precision.
This reflects our preference for conservative labeled evaluation: when a predicted question is counted as correct, it should reliably express the same semantic role as the gold question.
The resulting optimal threshold was \textbf{0.78}, yielding $F_{0.5}=0.90$, with precision $=0.96$ and recall $=0.70$ on the annotated sample.

\paragraph{Precision--recall tradeoff and error profile.}
The calibrated Semantic Match criterion exhibits very high precision, indicating that false positives are rare.
In contrast, recall is more moderate, implying that false negatives are substantially more common.
To better characterize these false negatives, we manually inspected 50 Hebrew cases in which predicted answers passed the IOU-based Argument Match stage but failed the Semantic Match threshold.
Half of these cases (25/50) were judged to be semantically acceptable by human inspection, typically corresponding to valid paraphrases or alternative but legitimate question formulations.
This analysis indicates that the Semantic Match metric is intentionally conservative and tends to underestimate true labeled performance rather than overestimate it.

\paragraph{Cross-lingual application.}
Although threshold calibration was performed using Hebrew data, we apply the same threshold uniformly across all target languages.
This choice is motivated by practical considerations and by the multilingual training of the underlying paraphrase model, under the assumption that cosine similarity scores are comparable across languages.
Importantly, this threshold is a hyperparameter of the \emph{evaluation metric} only: it does not affect the projection pipeline, model training, or deployment, and no ground-truth annotations are required at inference time.

\section{In-Context Prompt for QA Generation}
\label{app:icl_prompt}

Each model received few-shot demonstrations followed by the target sentence, with the predicate highlighted in bold.  
Two prompt templates were used: one for verbal predicates and one for nominal predicates.  
The structure and content were identical across all languages, with each model receiving the full prompt in its own language (Hebrew, Russian, or French).  
For clarity, the exact English versions of both prompts are shown below.

\paragraph{Verbal Predicate Prompt}
\begin{quote}
For a sentence with the predicate highlighted, create all questions and answers where the answers are found within the sentence.  
The answers must be a continuous fragment of the sentence, and the question must use the predicate as the main verb on which the question is asked.

For example: for the sentence:  
“Zeev Revach and Hanna Laslo are well-known and beloved comedians with great energy, who **performed** last weekend”

The questions and answers for the predicate “performed” are: \\ 
Who performed? → “Zeev Revach and Hanna Laslo are well-known and beloved comedians with great energy”  \\
When did someone perform? → “last weekend”

For this sentence and the predicate “**held**”:  
“These people **held** high and important positions in politics, administration, and business”

The questions and answers are:  \\
Who held? → “These people”  \\
Where did someone hold positions? → “in politics, administration, and business”

Here is a sentence with the predicate highlighted:  
“<sentence with **predicate**>”  
Please generate all questions and answers where the answers are found within the sentence.  
The answers must be a continuous fragment of the sentence, and the question must contain the predicate.
\end{quote}

\paragraph{Nominal Predicate Prompt}
\begin{quote}
For a sentence with the predicate highlighted, create all questions and answers where the answers are found within the sentence.  
The answers must be a continuous fragment of the sentence, and the question must use the predicate as an action noun, turning it into a verb, and ask a question on that verb.

For example: for the sentence:  
“We see that **the understanding** by Euler of the algorithm as a synonym for a method of solving a problem is already very close to the modern one.”

The questions and answers for the predicate “understanding” are:  \\
What is understood? → “the algorithm”  \\
How is something understood? → “as a synonym for a method of solving a problem”  \\
Who understands something? → “by Euler” \\ 
How does someone understand something? → “very close to the modern one” 

For the sentence:  
“The division commander Moshe Peled contested this and proposed instead to conduct an **attack** in the south of the Golan Heights.”

The questions and answers for the predicate “attack” are:  \\
Who can attack somewhere? → “The division commander Moshe Peled”  \\
Where can someone attack? → “in the south of the Golan Heights” 

For this sentence and the predicate “**appointment**”:  
“In the Supreme Court, judges serve by **appointment** permanently (until age 70), with the president of the Supreme Court at their head.”

The questions and answers are:  \\
Where was someone appointed? → “In the Supreme Court”  \\
Who was appointed? → “judges by”  \\
What appointment? → “permanently (until age 70)”

Here is a sentence with the predicate highlighted:  
“<sentence with **predicate**>”  
Please generate all questions and answers where the answers are found within the sentence.  
The answers must be a continuous fragment of the sentence, and the question must contain the predicate.
\end{quote}

\section{LoRA Adapter Configurations}
\label{appendix:finetuning}

All models were adapted using Low-Rank Adaptation (LoRA), with 4-bit NF4 quantization, linear adapter layers, dropout of 0.05, and gradient checkpointing. Optimization was performed using the AdamW optimizer with a linear learning rate schedule. Below we detail the adapter configurations and model identifiers used per language.

\paragraph{Hebrew.}
We trained LoRA adapters for \textbf{dicta-il/dictalm2.0-instruct}\footnote{\url{https://huggingface.co/dicta-il/dictalm2.0-instruct}}, a 7B Hebrew LLaMA 2 model, using:
\begin{itemize}[noitemsep, topsep=0pt]
\item Rank: 16, LoRA alpha: 64
\item Epochs: 25
\end{itemize}

As a multilingual baseline, we similarly trained LoRA adapters for \textbf{meta-llama/Meta-Llama-3-8B}\footnote{\url{https://huggingface.co/meta-llama/Meta-Llama-3-8B}} on the same Hebrew data and with identical hyperparameters.

\paragraph{Russian.}
We trained LoRA adapters for \textbf{sambanovasystems/SambaLingo-Russian-Base}\footnote{\url{https://huggingface.co/sambanovasystems/SambaLingo-Russian-Base}}, a 7B Russian-only model, with:
\begin{itemize}[noitemsep, topsep=0pt]
\item Rank: 8, LoRA alpha: 32
\item Epochs: 15
\end{itemize}

LoRA adapters were also trained for \textbf{meta-llama/Meta-Llama-3-8B}\footnotemark[2] on the Russian dataset using:
\begin{itemize}[noitemsep, topsep=0pt]
\item Rank: 16, LoRA alpha: 64
\item Epochs: 25
\end{itemize}

\paragraph{French.}
We trained LoRA adapters for \textbf{OpenLLM-France/Claire-7B-FR-Instruct-0.1}\footnote{\url{https://huggingface.co/OpenLLM-France/Claire-7B-FR-Instruct-0.1}}, a 7B French-only model, with:
\begin{itemize}[noitemsep, topsep=0pt]
\item Rank: 16, LoRA alpha: 64
\item Epochs: 10, Batch size: 64
\end{itemize}

Adapters were also trained for \textbf{meta-llama/Meta-Llama-3-8B}\footnotemark[2] on French using the same LoRA parameters with batch size 32.

\section{Manual Error Analysis of Semantic Evaluation}
\label{appendix:error_analysis}

This appendix provides a detailed analysis of Semantic Match failures referenced in the main paper.
We examine cases in which predicted answers passed the IOU threshold ($\geq 0.5$) but failed the semantic similarity check, focusing on error typology.

We manually analyzed 50 such cases from the Hebrew test set and assigned each predicted question to one of the following categories:
\begin{itemize}[noitemsep, topsep=0pt]
\item \textbf{M (Paraphrase Model Error):} The predicted question is semantically equivalent to the gold question, but the automatic similarity model fails to detect the match.
\item \textbf{V (Valid Alternate Question):} The predicted question is valid for the predicted answer, but targets a different argument than the gold annotation, typically due to minor span differences.
\item \textbf{P (Predicate Error):} The predicted question does not correctly correspond to the intended predicate.
\item \textbf{R (Role Labeling Error):} The question targets an incorrect semantic role.
\end{itemize}

The analysis shows that 50\% of the examined cases are semantically acceptable (30\% paraphrase model errors and 20\% valid alternate questions).
The remaining cases correspond to predicate realization errors (30\%) or role labeling mismatches (20\%).

Overall, this breakdown clarifies the sources of Semantic Match failures and complements the robustness analyses reported in the main text.


\end{document}